# Image Quality Assessment for Performance Evaluation of Focus Measure Operators

FARIDA MEMON*, MUKHTIAR ALI UNAR**, AND SHEERAZ MEMON***

## ABSTRACT

**This paper presents the performance evaluation of eight focus measure operators namely Image CURV (Curvature), GRAE (Gradient Energy), HISE (Histogram Entropy), LAPM (Modified Laplacian), LAPV (Variance of Laplacian), LAPD (Diagonal Laplacian), LAP3 (Laplacian in 3D Window) and WAVS (Sum of Wavelet Coefficients). Statistical matrics such as MSE (Mean Squared Error), PNSR (Peak Signal to Noise Ratio), SC (Structural Content), NCC (Normalized Cross Correlation), MD (Maximum Difference) and NAE (Normalized Absolute Error) are used to evaluate stated focus measures in this research. . FR (Full Reference) method of the image quality assessment is utilized in this paper. Results indicate that LAPD method is comparatively better than other seven focus operators at typical imaging conditions.**

**Key Word:** **Image Quality Evaluation, Focus Measure Operators, Shape From Focus, Image Processing.**

## 1. INTRODUCTION

In Computer Vision, reconstructing a three dimensional structure of an objet from a set of images is an important area of research. In SFF (Shape From Focus) method, a set of image sequence is taken at various focus settings and depth of the object is estimated by examining the best focused points taken from image set. SFF method has been effectively used in many industrial applications viz. microelectronics, industrial inspection, medical diagnostics, 3D cameras, and comparison of polymers [1-5].

The performance of this method depends upon the accuracy of the focus measure operators. This work presents the analysis of eight focus measure operators based on Image Quality Metrics. Image Quality assessment methods are generally categorized as: FR methods and NR (No Reference) methods. The quality of an image is measured in comparison with reference image in FR methods while NR methods do not use a reference image. The image quality metrics considered and implemented here fall in the FR category.

An extensive set of experiments is conducted using synthetic image sequences. Image Quality Assessment parameters such as MSE , PSNR , SC, NCC, MD, and NAE are carried out for comparing eight focus operators performance.

* Ph.D. Scholar, Institute of Information & Communication Technologies, Mehran University of Engineering & Technology, Jamshoro.
** Meritorious Professor, Department of Computer Systems Engineering, Mehran University of Engineering & Technology, Jamshoro.
*** Associate Professor, Department of Computer Systems Engineering, Mehran University of Engineering & Technology, Jamshoro.

## 2. SHAPE FROM FOCUS

Recovering the 3D shape of a scene from its 2D images is a central issue in the field of computer vision [6-8]. Based on optical reflective model, three dimensional shape recovery algorithms can be categorized as active and passive techniques. Projection of light rays is employed in active techniques whereas passive techniques are basically based on capturing of reflection of light rays. SFF is a passive method used to recover the 3D shape of the object. It is a well-known in the model of shape from X, where X represents the clue used to conclude the shape as, shading, stereo, motion, focus and de-focus.

In SFF technique, a sequence of an image is obtained at different focus positions. This can be noted that with limited depth of field, images obtained from lens contain both the areas in and out of focus. A quantity known as Focus measure is applied, to calculate the focus value of all pixels in the image sequence and a focus volume is achieved. By focus maximization, depth map is taken out from the focus volume. Thus, depth value of the certain pixel is calculated from the focus setting of the selected image. The relationship between the depth of object and lens focal length is given as:

$$\frac{1}{f} = \frac{1}{u} + \frac{1}{v} \qquad (1)$$

where *f*, *u* and *v* represent the lens focal length, depth of the object and distance between lens and image plane respectively. SFF technique can be employed by making a change in either of parameter *f*, *u*, *v* or any combination of them.

### 2.1 Focus Measure Operators

A Focus measure is stated as a quantity used to locally calculate the pixel sharpness [9].To calculate the focused point from set of images, several focus operators are proposed by different researchers. Focus measure operators can be grouped as [10]:

***Gradient-Based Operators:*** This group of focus measures use first derivative of the image or gradient to measure the focus level.

***Laplacian-Based Operators:*** These operators are similar to the earlier group. These focus measures are based on second derivative of an image.

***Statistics-Based Operators:*** To compute the focus level, this group of operators employ various statistical parameters of image as texture descriptors.

***Wavelet-Based Operators:*** These operators are based on the discrete wavelet transform coefficients. Thus these coefficients are used to compute the focus amount of an image.

***DCT-Based Operators:*** To calculate the degree of focus, this group of operators utilize the discrete cosine transform coefficients.

***Miscellaneous Operators:*** These operators do not belong to any of the previous five groups.

#### 2.1.1 Previous Research

Several types of focus measure operators are proposed by the different researchers. Some of these are reviewed as under.

Subbarao, et. al. [11] have proposed the numerous focus measure operators. These focus measures are based on energy of image gradient, image gray level variance, and energy of image Laplacian. For accurate shape recovery, Laplacian is the most suitable operator. In [11], second derivative of the image is utilized in calculation of a focus operator. Addition of squared 2nd derivative is used in calculation of ML (Modified Laplacian) operator. To increase the robustness of images with poor texture, Nayar, et. al. [12] have proposed a focus measure as sum of ML values around a small local window having dimensions of 5x5.

Helmli, et. al. [13] have presented typical algorithms for discovering sharp regions of image. They have proposed focus measure namely Mean method which is calculated as ratio of mean gray value to the center gray value in the small window. Approximation of a degree four polynomial is used in Point Focus Measure. 3D gradient used as a focus operator is proposed by Ahmad [14]. The focus measure proposed by Thelen, et. al. [15] includes diagonal neighbors of the image to calculate the sum of ML values of the image. Wee and Paramesran [16] has proposed the use of eigenvalues to measure the sharper image regions.

Different transforms are now widely used in image processing systems for focus measure calculation. In DCT (Discrete Cosine Transform) area, AC coefficients of corresponding image are computed for variance intensity calculation. Maximum value of variance represents that the image is well focused and it can be utilized as a focus measure. Shen and Chen [17] have taken AC and DC coefficients of a certain image and the ratio between both coefficients is described as a focus measure.

The Wavelet Transforms are also utilized as focus measure operators. Xie, et. al. [18] have utilized coefficients of wavelet transform as a focus measure.

Based on this literature review, eight algorithms are analyzed as mentioned in Section 4.2.

## 3. IMAGE QUALITY METRICS

In the development of image processing algorithms, IQM (Image Quality Measurement) plays an important role. To evaluate the performance of processed image, IQM can be utilized. Image Quality is defined as a characteristic of an image that measures the processed image degradation by comparing to an ideal image.

Humans are usually the observers and users of majority of imaging systems, hence image quality assessment by subjective method is considered to be the reliable method. However in real-time applications, use of subjective method is limited because of its complication and implementation difficulty. Therefore, objective methods are more widely used for image quality assessment in recent years.

In this work we consider several image quality metrics and analyze their statistical behavior for eight focus measures of SFF method.

### 3.1 Mean Squared Error

MSE is a very simple and common distortion measure. MSE between the reference image and processed image with a size of a ($m$x$n$) is expressed as follows:

$$MSE = \frac{1}{mn}\sum_{i=1}^{m}\sum_{j=1}^{n}(A_{ij} - B_{ij})^2 \quad (2)$$

whereand are the image pixel value of reference image and processed image respectively .

The value of MSE measures the difference between a processed image and reference image. The smaller value of the MSE represents the better result.

### 3.2 Peak Signal to Noise Ratio

For the measurement of reconstruction quality, PSNR (Peak Signal to Noise Ratio) is one of the most extensively used metric parameter. It describes the ratio of the maximum possible power of a signal to the power of corrupting noise and is normally represented in decibel scale. PSNR can be expressed as follows:

$$PSNR\ (db) = 10\ \log \frac{255^2}{MSE} \quad (3)$$

A higher value of PSNR specifies the reconstruction of higher quality.

### 3.3 Structural Content

This quality metric is expressed as follows:

$$SC = \frac{\sum_{i=1}^{m}\sum_{j=1}^{n}(A_{ij})^2}{\sum_{i=1}^{m}\sum_{j=1}^{n}(B_{ij})^2} \quad (4)$$

A higher value of SC (Structural Content) shows that image is of poor quality.

### 3.4 Normalized Cross Correlation

NCC (Normalized Cross Correlation) measure shows the comparison of the processed image and reference image. NCC is expressed as follows:

$$NCC = \sum_{i=1}^{m}\sum_{i=1}^{n}\frac{(A_{ij} \times B_{ij})}{A^2{}_{ij}} \quad (5)$$

### 3.5 Maximum Difference

MD (Maximum Difference) provides the maximum of the error signal (i.e. difference between the processed and reference image). MD is defined as follows:

$$MD = Max\left(\left|A_{ij} - B_{ij}\right|\right) \quad (6)$$

$i=1,2\ldots\ldots\ldots m, j = 1,2\ldots\ldots..n$

The higher the value of the maximum difference, the poorer the quality of the image.

### 3.6 Normalized Absolute Error

This quality measure can be expressed as follows.

$$NAE = \frac{\sum_{i=1}^{m}\sum_{j=1}^{n}\left(\left|A_{ij} - B_{ij}\right|\right)}{\sum_{i=1}^{m}\sum_{j=1}^{n}(A_{ij})} \quad (7)$$

A higher NAE value shows that image is of poor quality,

### 3.7 Average Difference

AD (Average Difference) provides the average of change concerning the processed and reference image. AD can be expressed as follows:

$$AD = \frac{1}{mn}\sum_{i=1}^{m}\sum_{i=1}^{n}\left[A(i, j) - B(i, j)\right] \quad (8)$$

Ideally it should be zero.

## 4 RESULTS

### 4.1 Test Images

In our study, experiments were performed on synthetic images of a cone. To examine the performance of different focus measures, several experiments were conducted using a set of 60 images of a simulated cone each having a dimensions of 302x302.

A MATALB program was initiated to produce the sequence of 60 images for a cone object. The details concerning the process are given in [19]. Simulated cone object was selected for the experimental work, as for a given ground truth depth map, verification of the results for such object is more easy. Sample images of a cone at various focus settings are shown in Fig. 1.

It is obvious that, in one image one part of the cone is well focused, whereas in other images, other parts of cone are well focused.

The objective was to acquire all in focus image with all parts focused and then to generate the 3D shape of the cone.

### 4.2 Experiments with Synthetic Images

For the performance evaluation of the focus measures, the existing methods are selected from different families described in Section 2.1. The details of mathematical description about these focus measure operators are given in [10].

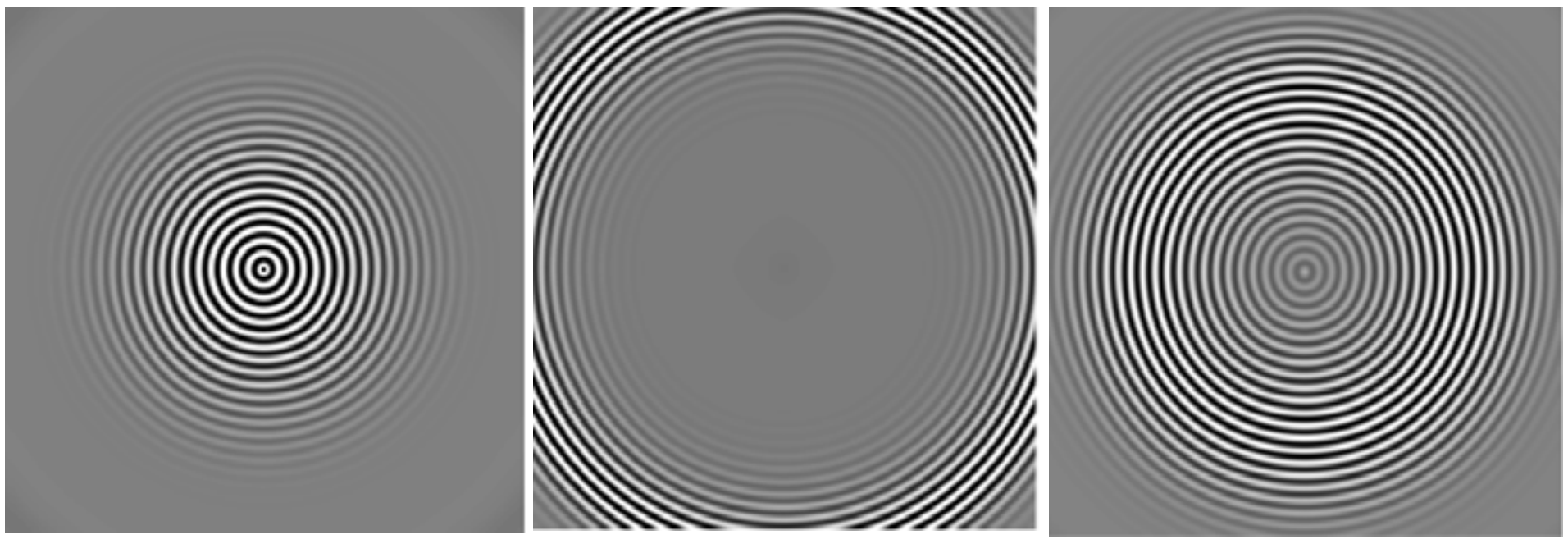

*FIG. 1. SAMPLE IMAGES OF A SIMULATED CONE*

For the application of shape recovery, Malik, et. al. [20] worked for the problem of determining the optimum window size of focus measures. They revealed that because of over smoothing effects, large size of window can produce inaccurate depth estimation. Conversely, as described in [21], small windows are more sensitive to noise and create the problem of image occlusion blur.

After extensive simulations, it was found that a suitable window size should be 7x7 and this window size was used in all investigations done in this study.

All the estimated focus measures were executed in MATLAB and experiments were conducted with a sequence of 60 images of simulated cone object as mentioned in Section 4.1.

Fig. 2 shows the experimental results, where Fig. 2(a) represents the ground truth images while Fig. 2(b-i) represent the simulation results. Fig. 2(i) shows the all-in focus images and Fig. 2(ii) shows the depth maps. All the results are achieved without adding of noise to the image sequence.

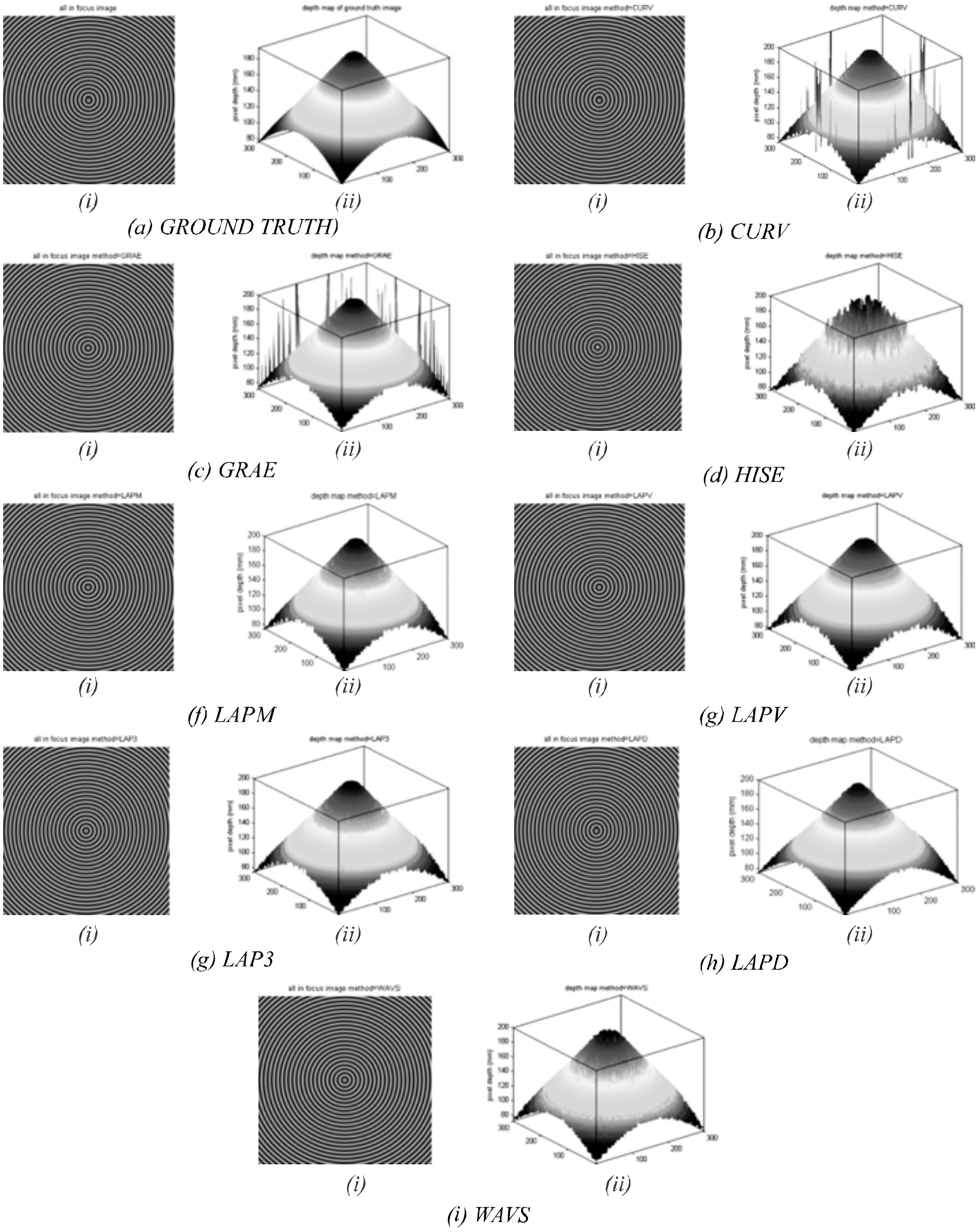

*FIG. 2. DEPTH MAPS AND ALL-IN FOCUS IMAGES FOR SIMULATED CONE*

From Fig 2, it is observed that Laplacian-based family operators perform well. Good results obtained with LAPM, LAPV and LAP3 operators, particularly more accurate depth map is achieved with LAPD operator. Whereas results of CURV, GRAE, HISE operators are considerably degraded and showing spikes in the depth maps. Furthermore, result of WAVS operator is clearly identifiable but shows the degraded central part of the cone.

From the experimental results, it is found that LAPD operator shows comparatively better performance than other seven.

### 4.3 Performance Analysis

In this study, statistical metrics namely MSE, PSNR, NCC, AD, SC, MD, and NAE were used to evaluate the performance of implemented focus measures. Results of statistically assessed focus measures are shown in Tables 1-2. Tables 1-2 compare the various statistical metrics of depth maps and all-in focus images in the absence of noise and also give the ideal values of these metrics. From Tables 1-2 it is clear that statistical metrics of Laplacian-based operators have more closed to the ideal values. Significantly better results are achieved with LAPD focus measure. For example MSE is a very simple and very common distortion measure. The value of MSE represents the difference between a processed and the reference image. The smaller the MSE, the better the result will be. Similarly for the measurement of reconstruction quality, PSNR is one of the most extensively used metric parameter. A higher PSNR value specifies the reconstruction of higher quality.

TABLE 1. COMPARISON OF FOCUS MEASURE METHODS WITH VARIOUS METRICS (DEPTH MAP)

| Method | MSE | PSNR | NCC | AD | SC | MD | NAE |
|---|---|---|---|---|---|---|---|
| Ideal Values | 0 | Inf | 1 | 0 | 1 | 0 | 0 |
| CURV | 23.6105 | 34.3998 | 0.9937 | 0.6101 | 1.0114 | 26.0196 | 0.0293 |
| GRAE | 26.0291 | 33.9762 | 0.9916 | 0.8768 | 1.0157 | 26.7141 | 0.0265 |
| HISE | 44.1025 | 31.6862 | 0.9976 | 0.0706 | 1.0024 | 25.3392 | 0.0411 |
| LAPM | 23.8293 | 34.3597 | 0.9940 | 0.5043 | 1.0109 | 9.8613 | 0.0337 |
| LAPV | 19.4382 | 35.2442 | 0.9905 | 1.0072 | 1.0182 | 9.3848 | 0.0299 |
| LAPD | 18.7301 | 35.4054 | 0.9941 | 0.6161 | 1.0111 | 9.4204 | 0.0293 |
| LAP3 | 19.1781 | 35.3027 | 0.9941 | 0.5991 | 1.0109 | 9.4204 | 0.0296 |
| WAVS | 31.2616 | 33.1807 | 0.9953 | 0.3511 | 1.0078 | 12.1735 | 0.0386 |

TABLE 2. COMPARISON OF FOCUS MEASURE METHODS WITH VARIOUS METRICS (ALL IN FOCUS)

| Ideal Values | 0 | Inf | 1 | 0 | 1 | 0 | 0 |
|---|---|---|---|---|---|---|---|
| CURV | 4.9967 | 41.1440 | 0.9990 | 0.1218 | 1.0018 | 15 | 0.0032 |
| GRAE | 16.4495 | 35.9693 | 0.9959 | 0.4973 | 1.0072 | 65 | 0.0051 |
| HISE | 25.7436 | 34.0241 | 0.9918 | 0.2492 | 1.0149 | 29 | 0.0257 |
| LAPM | 3.0770 | 43.2495 | 0.9955s | 0.6025 | 1.0072 | 15 | 0.0088 |
| LAPV | 1.7100 | 45.6463 | 0.9951 | 0.5391 | 1.0097 | 12 | 0.0052 |
| LAPD | 1.7719 | 45.8008 | 0.9963 | 0.4999 | 1.0089 | 10 | 0.0050 |
| LAP3 | 1.9243 | 45.2882 | 0.9955 | 0.5108 | 1.0089 | 12 | 0.0053 |
| WAVS | 7.2636 | 39.5193 | 0.9986 | 0.5263 | 1.0023 | 27 | 0.0152 |

Results given in Tables 1-2 indicate that as compared with other seven focus measures, LAPD has lower MSE value while PSNR value is higher.

## 5 CONCLUSION

In this work, several focus operators published in the literature are studied. To evaluate the performance of the focus measures, existing methods were selected from different families. The conducted experiments and results of various statistical measures have shown that LAPD operator gives more accurate 3D shape recovery. At typical imaging conditions (like without addition of noise), it shows comparatively better performance than other seven CURV, GRAE, HISE, LAPM, LAPV, LAP3, WAVS operators.

## ACKNOWLEDGEMENT

The first author acknowledges the financial suport through Institute of Information & Communication Technologies, Endowment Fund of Mehran University of Engineering & Technology, Jamshoro, Pakistan, to carryout this research. The second and third authors are her Supervisor and Co-Supervisor, respectively.